%% file: nns4ssr.tex
\documentclass{article}
\usepackage{spconf,amsmath,graphicx}
\usepackage{multirow}

\input{math_commands}

\input{macros}

\usepackage{hyperref}
\usepackage{url}
\usepackage{tikz,pgfplots}
\usepackage{array} 
\usepgfplotslibrary{external}
\usepackage{enumitem}
\usepackage{subcaption}
 \tikzsetexternalprefix{figs_tikz/}

\title{A Learning-Based Framework for Line-Spectra Super-resolution}
\name{Gautier Izacard$^{\star \ddagger}$ \qquad Brett
  Bernstein$^{\dagger}$\thanks{B.B. is supported by a MacCracken,
    and a Barkey and
    Yhap fellowship.} \qquad Carlos Fernandez-Granda$^{\star \dagger}$\thanks{C.F. is supported by NSF award DMS-1616340.}}
\address{$^{\star}$ Center for Data Science, NYU, New York, USA \\
			    $^{\ddagger}$ Ecole Polytechnique, Paris, France\\
			    $^{\dagger}$ Courant Institute of Mathematical Sciences, NYU, New York, USA}

\begin{document}
%
\maketitle
\begin{abstract}
We propose a learning-based approach for estimating the spectrum of a
multisinusoidal signal from a finite number of samples. A
neural-network is trained to approximate the spectra of such signals on simulated data. The proposed methodology is very flexible: adapting to different signal and noise models only requires modifying the training data accordingly. Numerical experiments show that the approach performs competitively with classical methods designed for additive Gaussian noise at a range of noise levels, and is also effective in the presence of impulsive noise.
\end{abstract}

\input{intro}

\input{related_work}

\input{methodology}
\input{numerical}

\input{conclusion}

\bibliographystyle{IEEEbib}
\bibliography{bibli.bib}

\end{document}

%% file: math_commands.tex
\usepackage{amsmath,amsfonts,bm}

\def\eqref#1{equation~\ref{#1}}

\def\1{\bm{1}}

\DeclareMathAlphabet{\mathsfit}{\encodingdefault}{\sfdefault}{m}{sl}
\SetMathAlphabet{\mathsfit}{bold}{\encodingdefault}{\sfdefault}{bx}{n}

\newcommand{\R}{\mathbb{R}}

\DeclareMathOperator{\sign}{sign}

%% file: macros.tex
\newcommand{\C}{\mathbb{C}}
\newcommand{\CC}{\mathbb{C}}

\newcommand{\RR}{\mathbb{R}}

\newcommand{\Ind}{\mathbf{1}}

\newcommand{\perm}{\mathfrak{S}}

\newcommand{\abs}[1]{\left| #1 \right|}

\newcommand{\brac}[1]{\left( #1 \right) }

\newcommand{\op}[1]{ \operatorname{#1} }

\newcommand{\normTwo}[1]{\left\| #1 \right\| _{2}}

%% file: intro.tex
\section{Introduction}
\subsection{Super-resolution of Spectral Lines}
Estimating the spectra of multisinusoidal signals from a finite number
of noisy samples is a fundamental problem in signal processing, with
applications in sonar, radar, communications, geophysics, speech
analysis, and other domains (see \cite{stoica1993list} for an
extensive list of references). Consider a signal $S:\RR\to\CC$ given by
\begin{align}
\label{eq:model}
S(t) := \sum_{j=1}^{m} a_j \exp (i 2 \pi f_j t),
\end{align} 
where $a_j \in \C$  denotes the amplitude of the $j$th sinusoidal component. The Fourier transform of such signals is a superposition of Dirac deltas, or \emph{spectral lines}, located at the frequencies $f_1$, $f_2$, \ldots, $f_m \in \R$. Our goal is to estimate these frequencies from a \emph{finite} number of noisy samples obtained at the Nyquist rate. Assuming (without loss of generality) that $0 \leq f_j \leq 1$, $1 \leq j \leq m$, so that the Nyquist rate is equal to one, the data are given by
\begin{align}
\label{eq:data}
y_k := S(k) + z_k, \qquad 1 \leq k \leq n,
\end{align}
where $z_k \in \C$ is an additive perturbation. The line-spectra estimation problem is often referred to as \emph{spectral super-resolution} because truncating the signal in time is equivalent to convolving the line spectra with a blurring sinc kernel of width $2/n$. As a result, the spectral resolution of the data is limited by the number of available samples even in the absence of noise. 

\subsection{Contributions}

Inspired by recent advances in deep learning~\cite{lecun2015deep}, we
present a learning-based methodology to tackle the line-spectra
estimation problem that involves calibrating a deep neural network using simulated data. Training the neural network is costly but can be carried out offline. However, once the model is trained, processing new signals by applying the forward propagation of the neural network is significantly faster than applying variational techniques. An insight underlying the design of the network is that estimating frequency locations directly is less effective than generating a smoothed estimate of the spectrum, as demonstrated in Section~\ref{sec:direct}. In Section~\ref{sec:compare} the approach is shown to perform competitively with classical methods in a range of signal-to-noise ratios (SNRs), matching the robustness of linear nonparametric estimators at low SNR, as well as the accuracy of parametric estimators at high SNR. Finally, Section~\ref{sec:sparse} illustrates the flexibility of our framework with an application to line-spectra estimation in the presence of sparse noise.

%% file: related_work.tex
\section{State of the Art and Related Work}
\label{sec:related_work}

The simplest method to estimate line spectra is to window the data and then compute its Fourier transform. This technique, known as the periodogram, provides a linear nonparametric estimate composed of a superposition of sinc kernels of width $1/2n$ centered at the position of the line spectra. The interference between these kernels complicates locating the line spectra precisely. As a result, the periodogram does not yield an exact estimate of the spectrum, even in the absence of noise. However, when the SNR is low, the windowed periodogram is an effective estimator (see Chapter 2 in \cite{Stoica:2005wf}).  

In contrast to the periodogram, Prony's method (see \cite{deProny:tg}, as well as \cite{fri} for a modern exposition) is guaranteed to recover line spectra exactly from noiseless measurements. Parametric methods based on Prony's method are very popular and highly effective at high and medium SNRs. They include techniques based on matrix pencils~\cite{hua1990matrix}, and subspace methods such as MUSIC (MUltiple SIgnal Classification)~\cite{music1,music2}. These approaches are significantly more computationally heavy than the periodogram because they require computing eigendecompositions of matrices built from the samples.

Recently, variational techniques based on sparse recovery have been proposed for the line-spectra estimation problem~\cite{bhaskar2013atomic,candes2014towards}. This approach is computationally intensive, as it requires solving a semidefinite program or a convex program involving a large dictionary of discretized sinusoids. An important advantage is that it can easily be adapted to deal with missing data~\cite{cs_offgrid} and outliers \cite{fernandez2016demixing}. 

Finally, from a methodological point of view, our work is related to recent neural-network based approaches to sparse recovery \cite{xin2016maximal,he2017bayesian}, point-source deconvolution \cite{boyd2018deeploco}, and acoustic source localization \cite{adavanne2017direction,xiao2015learning,chakrabarty2017broadband}.

%% file: methodology.tex
\section{Methodology}
\label{sec:methodology}

We propose to perform line-spectra super-resolution using a deep neural network, which we call pseudo-spectrum net (PSnet). The input to the network are data generated according to the model in~\eqref{eq:model}. The output of the network is an approximation to the spectrum, called a \emph{pseudo-spectrum}. Numerical experiments reported in Section~\ref{sec:direct} show that generating such an approximation is more effective than trying to train a network to output the frequency locations of the spectral lines directly. A similar phenomenon has been observed when calibrating small deep-learning models to approximate larger networks: approximating the softmax outputs of the larger model produces better results than trying to fit its discrete predictions~\cite{hinton2015distilling}.

We define the pseudo-spectrum $P_{f}: \R \rightarrow \R$ as the convolution of the spectral lines of the signal with a kernel $K: \R \rightarrow \R$:
\begin{align}
  \label{eq:pseudo}
P_{f}\brac{u} & := \sum_{i=1}^{m} K\brac{u-f_j}, 
\end{align}
where $f$ is a vector containing $m$ frequencies. In all our numerical
experiments, $K$ is a triangular kernel. The PSnet is a neural network
$\op{PS}_{w}: \C^n \rightarrow \R^g$ parametrized by linear weights
$w$ that outputs a fine discretization of the pseudospectrum on a grid
of size $g$.  The network treats complex numbers as pairs of real
numbers, which makes it possible to leverage standard optimization packages to train it.
The weights are calibrated by using the
Adam optimizer~\cite{kingma2014adam} to minimize the approximation error between the output
of the network and the discretized pseudospectrum on a training set with $N$ examples:
\begin{align}
\small{
w := \min_{\tilde{w} } \sum_{l=1}^{N} \normTwo{P_{f^{[l]}} - F_{\tilde{w} }(y^{[l]}) }^2.}
\end{align}
The $l$th training example is given by
\begin{align}
 y^{[l]}_k &:= \sum_{j=1}^{m} a^{[l]}_j \exp (i 2 \pi f^{[l]}_j k) + z^{[l]}_k, \; 1\leq k \leq n, \, 1 \leq l \leq N,  \notag
\end{align}
where the frequencies $f^{[l]}$, the amplitudes $a^{[l]}$ and the noise $z^{[l]}$ are sampled from predefined distributions. Once the PSnet is calibrated, the position of the line spectra for a new vector of noisy data $y$ can be estimated by locating the peaks of the corresponding estimated pseudospectrum $\op{PS}_{w }(y)$. 

The architecture of the PSnet consists of a linear layer followed by several convolutional layers and a final linear layer.  The layers are separated by rectified linear units (ReLUs), a standard non-linearity in deep learning, and include batch normalization~\cite{ioffe2015batch}. Intuitively, the first layer maps the data to a frequency representation (one can check that the rows of the corresponding matrices are sinusoidal). In the frequency domain, the contribution of each spectral line to the data is concentrated around it and displays translation invariance: shifting the line just shifts its corresponding component in the data. This motivates using convolutional layers, which consist of localized filters that are convolved with the input, to build the rest of the network. In computer vision, convolutional layers are a fundamental tool for exploiting translation invariance~\cite{lecun1998gradient}. 



%% file: numerical.tex
\section{Numerical Experiments}

\input{experimental_design}

\input{direct}


\input{compare}


%

\input{sparse}

%% file: experimental_design.tex

In this section, we provide numerical evidence that the PSnet
generalizes effectively on test data not present in the training set
used to calibrate the model. In all experiments, we train the network
on a variable number of frequencies, i.e., the number of spectral
lines is not fixed in the training set.



\subsection{Experimental Design}
\label{sec:experimental_design}
In our experiments, the training and test sets are generated by sampling the frequency locations, amplitudes and noise in the measurement model of \eqref{eq:data} independently at random. The coefficients are given by $a_j:=(0.1+|w_j|)e^{ i \theta_j}$, $j=1,\ldots,m$, where $w_j$ is standard Gaussian distribution and $\theta_j$ is uniform in $[0,2\pi]$. In all sections except \ref{sec:sparse} $z_l$ is standard Gaussian noise scaled to ensure a given SNR. 

In order to design an appropriate distribution for the frequencies of the spectral lines, it is necessary to take into account that the minimum separation $\Delta:= \min_{j \neq j'} \abs{f_{j}-f_{j'}}$ between them determines whether the problem is well posed. At minimum separations below $\Delta := 1/n$ the problem is severely ill posed, in the sense that estimating the amplitudes requires solving a linear system that is very ill conditioned even if the true frequencies are known~\cite{moitra_superres}. To ensure that the training set contains well posed instances the inter-frequency separations of each signal are given by
$f_{j+1}-f_j=\brac{ \sign \brac{u_j} \Delta_{\min} + u_j} \bmod 1$ for $j=1,\ldots,m-1$ where the $u_j$ are
i.i.d.~drawn from a centered Gaussian distribution with standard deviation $2.5/n$ and $\Delta_{\min} \geq 1/n$.  Finally, $n$ is fixed to 50 and the number of
frequencies $m$ is chosen uniformly between $1$ and $10$.

In order to produce an estimate of the frequencies from the pseudospectrum generated by the network we locate the highest $m$ peaks, where for each signal the number m of frequencies is assumed known. The same assumption is needed to extract frequency estimates from traditional techniques such as the periodogram and MUSIC. Estimating the number of frequencies automatically is an important problem in itself that we did not consider in this paper.
To measure recovery accuracy we use two metrics: false-negative rate, and matched distance. The false negative rate is defined by
{\small
$$\op{FN}(f,\hat{f}) = \frac{1}{m}\sum_{j=1}^{m} \Ind\brac{
\{\hat{f}_1,\ldots,\hat{f}_{m}\} \cap [f_j-\frac{1}{2n},f_j+\frac{1}{2n}]=\emptyset}.$$}In words, a false negative occurs when there is no estimated frequency
that is closer than $\frac{1}{2n}$ to a true frequency. The matched distance between the true frequencies and the estimate, denoted by $\hat{f}_1,\ldots,\hat{f}_{m}$,
is given by
$$\op{MD}(f,\hat{f}) = n \cdot \op{avg}\{ \op{avg}_j\min_k|f_j-\hat{f}_k|, \op{avg}_k\min_j|f_j-\hat{f}_k|\}.$$
In words, we match each frequency with its closest counterpart,
and record the average error, normalized by $1/n$. To remove the influence of large errors that are accounted for by the false negative rate, we only average over frequencies where the closest counterpart is within $\frac{1}{2n}$.

%% file: direct.tex
\begin{table}[]
\small
\begin{tabular}{l|cc|cc|cc}
\hline
SNR & \multicolumn{2}{c|}{1} & \multicolumn{2}{c|}{100} & \multicolumn{2}{c}{10000} \\ \cline{2-7} 
                     & FN      & MD      & FN       &  MD      & FN       & MD       \\ \hline
Pair.             & 52.7\%     & 0.478     & 19.4\%      & 0.286     & 23.4\%      & 0.331     \\
DL             & 38.3\%     & 0.433     & 13.1\%      & 0.229    & 21.1\%      & 0.294      \\
PS           & 15.8\%     & 0.137     & 11.1\%       & 0.099     & 15.1\%       & 0.122     \\  \hline
\end{tabular}
\caption{Comparison of the average false-negative rate (FN) and matched-distance error (MD) normalized by $1/n$ for the experiment described in Section~\ref{sec:direct}. The neural networks trained using the pairing (Pair.) and the DeepLoco (DL) losses are outperformed by the network trained to output a pseudo-spectrum (PS) over the three different SNRs.
}
\label{table:direct}
\end{table}

\subsection{Comparison to Direct Estimation of Frequencies}
\label{sec:direct}
Our proposed methodology is based on producing a pseudo-spectrum from which to estimate spectral-line locations. In this section we compare this choice to the alternative approach of training a neural network to directly output the frequency estimates ${\hat{f}_1, \ldots , \hat{f}_m}$. This requires a careful choice of the training loss used to calibrate the network. A natural approach is to associate each frequency of the signal to an element of the output using the minimal pairing distance over all possible permutations $\perm$,
\begin{align}
\small
\min_{\sigma \in \perm} \sum_{j=1}^{m} \abs{f_j-\hat{f}_{\sigma \brac{j}}}^2.
\end{align}  
Recent work on point-source deconvolution~\cite{boyd2018deeploco} introduces an alternative loss, where the distance between the estimated and the true frequencies is computed after smoothing with a kernel (e.g. a Laplacian or Gaussian kernel). This approach is closer to pseudo-spectrum estimation; it computes a pseudo-spectrum that is parametrized by the estimated frequencies. In contrast, our methodology produces a nonparametric estimate of the pseudo-spectrum.

To compare direct frequency estimation with our proposed methodology we use the same architecture to perform direct estimation and to estimate a pseudo-spectrum. We fix the architecture to be a fully connected network with 9 hidden layers, the first of which contains $5000$ neurons and the rest of which contain 500 neurons. Empirically, this seems to yield the best results for the direct-estimation losses. By adding a last linear layer with an output of dimension $m$, the network can be trained to produce frequency estimates using the minimal pairing distance and the DeepLoco loss. By adding a last layer with an output of dimension $g:=10^3$ it can be trained using our methodology to produce an estimate of the pseudo-spectrum. 

\input{bfigs2}
\input{compare_table}
A complication that arises when performing direct estimation is how to output a variable number of estimated frequencies. Our approach does not suffer from this problem; a varying number of spectral lines simply results in a different number of peaks in the estimated pseudo-spectrum. However, here we consider a simple case where $m$ is fixed and equal to two ($m:=2$). Table~\ref{table:direct} compares the performance of these three different options at three different SNRs ($1$, $100$, and $10^4$). At each SNR the training and test sets contain $10^4$ signals generated as described in Section~\ref{sec:experimental_design} with $\Delta_{\min}:=2/n$. Our results suggest that generating a pseudo-spectrum significantly outperforms direct estimation of frequencies. Designing an architecture to produce accurate frequency estimates directly is an interesting direction for future research.

%% file: bfigs2.tex
\begin{figure}
  \centering
  \includegraphics[scale=0.46]{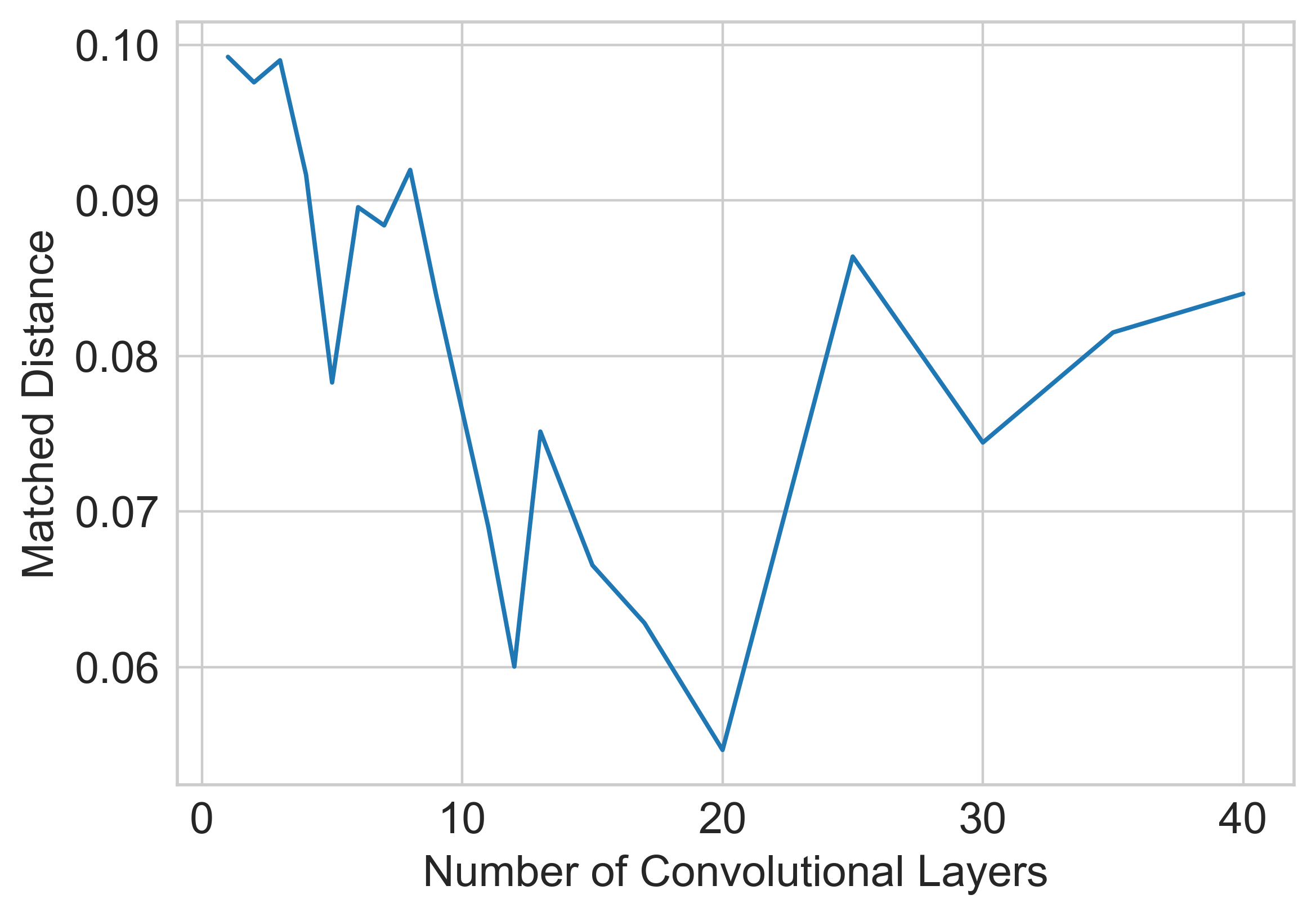}
  \caption{Plot of validation set performance as a function of network depth at
    SNR $10^4$.  Each layer of the network consists of 8 circular convolution
    filters (with filter size 3), batch normalization, and a ReLU
    non-linearity as described in Section~\ref{sec:methodology}.
  }
  \label{fig:depth}
\end{figure}

%% file: compare_table.tex
\begin{table*}[ht]
  \centering
  \small
\begin{tabular}{l|cc|cc|cc|cc}
\hline
SNR & \multicolumn{2}{c|}{1} & \multicolumn{2}{c|}{100} & \multicolumn{2}{c|}{10000} & \multicolumn{2}{c}{Blind}\\ \cline{2-9} 
                     & FN     & MD      & FN       & MD    & FN   & MD  & FN       & MD       \\ \hline
MUSIC             & 38.04\%     & 0.144     & 8.79\%      & 0.054     & 3.02\%      & 0.087     & 4.95\% & 0.032\\
Periodogram      & 30.14\%     & 0.120     & 13.73\%      & 0.091    & 13.36\%      & 0.087      & 15.47\% & 0.089\\
PSnet           & 27.04\%     & 0.143     & 2.62\%       & 0.063    & 2.10\%       & 0.054     & 2.63\% & 0.061\\  \hline
\end{tabular}
\caption{Comparison of the average false-negative rate (FN) and matched-distance error (MD) normalized by $1/n$ for the experiments described in Section~\ref{sec:direct}.}
\label{table:compare}
\end{table*}

%% file: compare.tex
\subsection{Comparison to Traditional Methods}
\label{sec:compare}
In this section we compare the performance of the PSnet to two of the
main traditional methods for line-spectra estimation: the
periodogram~\cite{Stoica:2005wf}, and MUSIC~\cite{music1,music2} (see
Section~\ref{sec:related_work}). We train PSnets with the architecture
detailed in Section~\ref{sec:methodology} on data generated as
described in Section~\ref{sec:experimental_design} with $\Delta_{\min}:=1/n$ for a range of
SNRs. Figure~\ref{fig:depth} shows the performance of the network in
terms of matched distance error for different depths. For the
comparison with other methods, we set the number of convolutional
layers at 20, each with 8 filters (each of size 3), and the dimension of the initial linear layer to 100.

The results of the comparison are shown in Table~\ref{table:compare}. We calibrate a different PSnet on training data with an SNR of 1, 100 and $10^4$. In addition, we train a single PSnet for a \emph{blind}-noise scenario where the noise level is not known beforehand by varying the SNR of the signals in the training and test sets (the square root of the SNR is uniformly sampled between 1 and 100). The training set contains $2\cdot10^5$ signals in every case. In the high noise regime (SNR 1) the PSnet and the periodogram have similar performance, while MUSIC has a considerably larger false negative rate. In the lower noise regimes (SNR 100 and $10^4$) the PSnet and MUSIC outperform the periodogram, with the PSnet having the lowest false negative rate. In the blind-noise regime, the PSnet again outperforms both other methods in terms of false-negative rate and is competitive with MUSIC in matched-distance error.

%% file: sparse.tex
\subsection{Line-Spectra Estimation from Corrupted Data}
\label{sec:sparse}

A promising feature of learning-based methods is that they can easily incorporate prior assumptions on the measurements. In this section we consider the problem of performing line-spectra estimation when a subset of the data are completely corrupted, i.e., when the vector $z$ in \eqref{eq:data} is sparse. In particular, we consider a regime where the corruptions have a standard deviation on the same order as the amplitude of the sampled signal, so they produce significant perturbations while being challenging to detect. 

\begin{figure}
\begin{subfigure}{.25\textwidth}
  \centering FN\\
  \includegraphics[width=\linewidth]{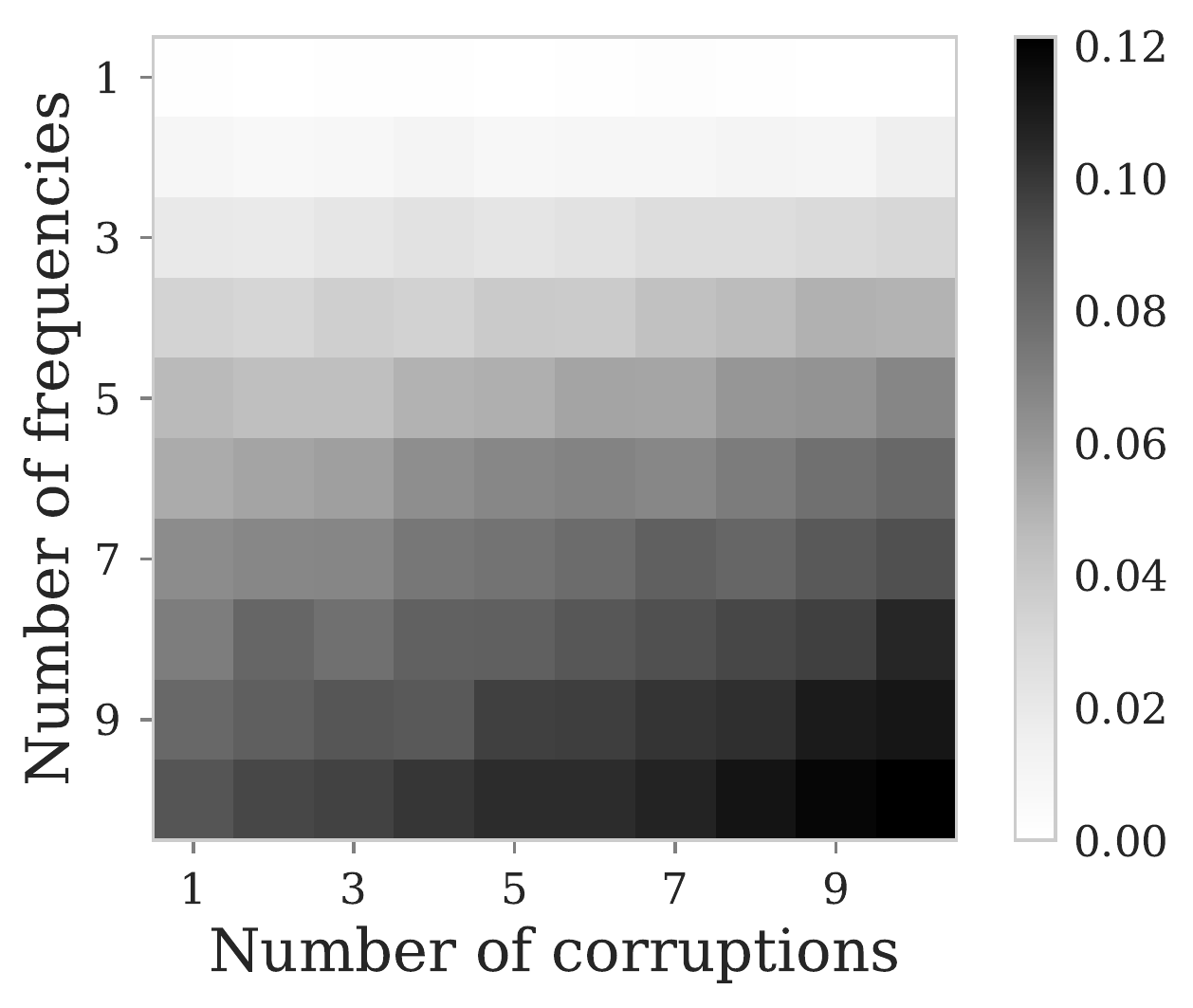}
\end{subfigure}%
\begin{subfigure}{.25\textwidth}
  \centering MD\\
  \includegraphics[width=\linewidth]{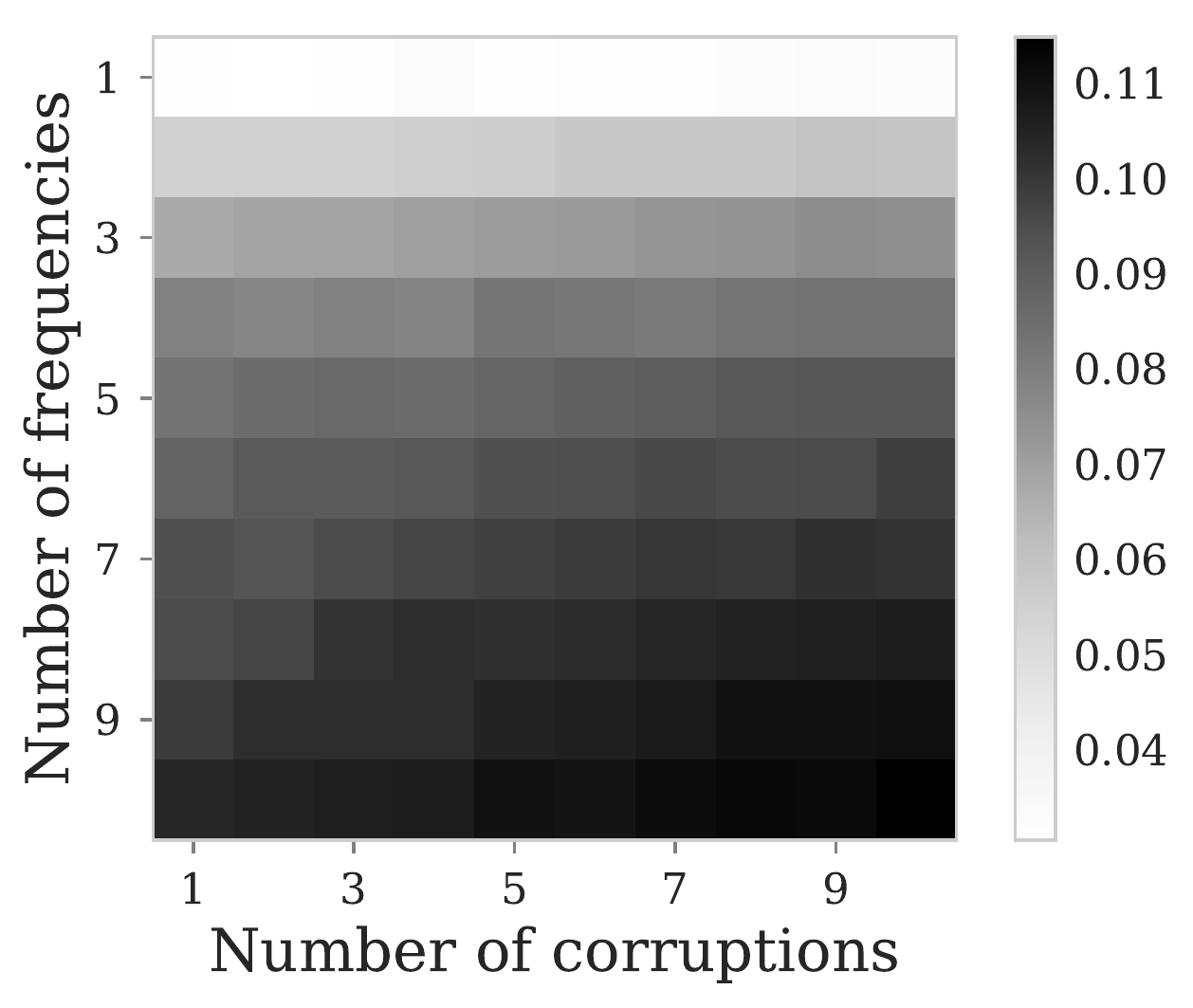}
\end{subfigure}
\caption{Comparison of the average false-negative rate (FN) and matched-distance error (MD) normalized by $1/n$ for the experiment described in Section~\ref{sec:sparse}.}
\label{fig:sparse}
\end{figure}

To evaluate the performance of our network we generate training and test sets with $2\cdot10^4$ examples, where each example has between 1 and 10 spectral lines. The data are simulated as described in Section~\ref{sec:experimental_design} with $\Delta_{\min}:=1/n$, except for the noise. The noise is set to have a support with fixed cardinality ranging from 1 to 10 (i.e., up to 20\% of the measurements). Its amplitude is i.i.d. Gaussian with a standard deviation equal to 1/2 (for reference the $\ell_2$ norm of the signal is normalized). The network architecture follows the description in Section~\ref{sec:methodology}; the dimensionality of the linear layer is 500, the number of convolutional layers is 20, with 8 filters of size 3 per layer. Figure~\ref{fig:sparse} quantifies the quality of the estimate in the presence of outliers for the two metrics. As expected the quality decreases as the number of outliers and frequencies increases.

%% file: conclusion.tex
\section{Conclusion and Future Work}


Our results suggest that learning-based methods are a promising avenue for tackling signal-processing problems such as line-spectra estimation. An important difference between these approaches and traditional methods is that the performance of learning-based methods depends on the probabilistic assumptions encoded in the training set. This is an attractive feature, as it makes it straightforward to adapt the approach to different signal and noise models. However, it also presents a crucial challenge for future research: understanding under what conditions training on simulated measurements ensures robust generalization to real data.  

%% file: nns4ssr.bbl
\begin{thebibliography}{10}

\bibitem{stoica1993list}
Petre Stoica,
\newblock ``List of references on spectral line analysis,''
\newblock {\em Signal Processing}, vol. 31, no. 3, pp. 329--340, 1993.

\bibitem{lecun2015deep}
Yann LeCun, Yoshua Bengio, and Geoffrey Hinton,
\newblock ``Deep learning,''
\newblock {\em Nature}, vol. 521, no. 7553, pp. 436, 2015.

\bibitem{Stoica:2005wf}
Petre Stoica and R.~L. Moses,
\newblock {\em {Spectral analysis of signals}},
\newblock Prentice Hall, Upper Saddle River, New Jersey, 1 edition, 2005.

\bibitem{deProny:tg}
Baron Gaspard Riche~de Prony,
\newblock ``{Essai {\'e}xperimental et analytique: sur les lois de la
  dilatabilit{\'e} de fluides {\'e}lastique et sur celles de la force expansive
  de la vapeur de l'alkool, {\`a} diff{\'e}rentes temp{\'e}ratures},''
\newblock {\em Journal de l'{\'E}cole Polytechnique}, vol. 1, no. 22, pp.
  24--76, 1795.

\bibitem{fri}
Martin Vetterli, Pina Marziliano, and Thierry Blu,
\newblock ``Sampling signals with finite rate of innovation,''
\newblock {\em IEEE Trans. on Signal Processing}, vol. 50, no. 6, pp.
  1417--1428, 2002.

\bibitem{hua1990matrix}
Y~Hua and T.K Sarkar,
\newblock ``{Matrix pencil method for estimating parameters of exponentially
  damped/undamped sinusoids in noise},''
\newblock {\em {IEEE} Trans. Acoust., Speech, Signal Process.}, vol. 38, no. 5,
  pp. 814--824, May 1990.

\bibitem{music1}
G.~Bienvenu,
\newblock ``Influence of the spatial coherence of the background noise on high
  resolution passive methods,''
\newblock in {\em Proceedings of the International Conference on Acoustics,
  Speech and Signal Processing}, 1979, vol.~4, pp. 306 -- 309.

\bibitem{music2}
R.~Schmidt,
\newblock ``Multiple emitter location and signal parameter estimation,''
\newblock {\em IEEE Trans. on Antennas and Propagation}, vol. 34, no. 3, pp.
  276 -- 280, 1986.

\bibitem{bhaskar2013atomic}
Badri~Narayan Bhaskar, Gongguo Tang, and Benjamin Recht,
\newblock ``Atomic norm denoising with applications to line spectral
  estimation,''
\newblock {\em IEEE Trans. on Signal Processing}, vol. 61, no. 23, pp.
  5987--5999, 2013.

\bibitem{candes2014towards}
Emmanuel~J Cand{\`e}s and Carlos Fernandez-Granda,
\newblock ``Towards a mathematical theory of super-resolution,''
\newblock {\em Communications on Pure and Applied Mathematics}, vol. 67, no. 6,
  pp. 906--956, 2014.

\bibitem{cs_offgrid}
G.T. Tang, B.~N. Bhaskar, P.~Shah, and B.~Recht,
\newblock ``Compressed sensing off the grid,''
\newblock {\em IEEE Trans. on Information Theory}, vol. 59, no. 11, pp.
  7465--7490, 2013.

\bibitem{fernandez2016demixing}
Carlos Fernandez-Granda, Gongguo Tang, Xiaodong Wang, and Le~Zheng,
\newblock ``Demixing sines and spikes: {R}obust spectral super-resolution in
  the presence of outliers,''
\newblock {\em Information and Inference}, vol. 7, no. 1, pp. 105--168, 2017.

\bibitem{xin2016maximal}
Bo~Xin, Yizhou Wang, Wen Gao, David Wipf, and Baoyuan Wang,
\newblock ``Maximal sparsity with deep networks?,''
\newblock in {\em Advances in Neural Information Processing Systems}, 2016, pp.
  4340--4348.

\bibitem{he2017bayesian}
Hao He, Bo~Xin, Satoshi Ikehata, and David Wipf,
\newblock ``From {B}ayesian sparsity to gated recurrent nets,''
\newblock in {\em Advances in Neural Information Processing Systems}, 2017, pp.
  5554--5564.

\bibitem{boyd2018deeploco}
Nicholas Boyd, Eric Jonas, Hazen~P Babcock, and Benjamin Recht,
\newblock ``Deep{L}oco: Fast 3{D} localization microscopy using neural
  networks,''
\newblock {\em BioRxiv}, p. 267096, 2018.

\bibitem{adavanne2017direction}
Sharath Adavanne, Archontis Politis, and Tuomas Virtanen,
\newblock ``Direction of arrival estimation for multiple sound sources using
  convolutional recurrent neural network,''
\newblock {\em arXiv preprint arXiv:1710.10059}, 2017.

\bibitem{xiao2015learning}
Xiong Xiao, Shengkui Zhao, Xionghu Zhong, Douglas~L Jones, Eng~Siong Chng, and
  Haizhou Li,
\newblock ``A learning-based approach to direction of arrival estimation in
  noisy and reverberant environments,''
\newblock in {\em Proceedings of the International Conference on Acoustics,
  Speech and Signal Processing}, 2015, pp. 2814--2818.

\bibitem{chakrabarty2017broadband}
Soumitro Chakrabarty and Emanu{\"e}l~AP Habets,
\newblock ``Broadband {DOA} estimation using convolutional neural networks
  trained with noise signals,''
\newblock in {\em Applications of Signal Processing to Audio and Acoustics
  (WASPAA)}. IEEE, 2017, pp. 136--140.

\bibitem{hinton2015distilling}
Geoffrey Hinton, Oriol Vinyals, and Jeff Dean,
\newblock ``Distilling the knowledge in a neural network,''
\newblock {\em arXiv preprint arXiv:1503.02531}, 2015.

\bibitem{kingma2014adam}
Diederik~P Kingma and Jimmy Ba,
\newblock ``Adam: A method for stochastic optimization,''
\newblock {\em arXiv preprint arXiv:1412.6980}, 2014.

\bibitem{ioffe2015batch}
Sergey Ioffe and Christian Szegedy,
\newblock ``Batch normalization: Accelerating deep network training by reducing
  internal covariate shift,''
\newblock {\em arXiv preprint arXiv:1502.03167}, 2015.

\bibitem{lecun1998gradient}
Yann LeCun, L{\'e}on Bottou, Yoshua Bengio, and Patrick Haffner,
\newblock ``Gradient-based learning applied to document recognition,''
\newblock {\em Proceedings of the IEEE}, vol. 86, no. 11, pp. 2278--2324, 1998.

\bibitem{moitra_superres}
Ankur Moitra,
\newblock ``Super-resolution, extremal functions and the condition number of
  {V}andermonde matrices,''
\newblock in {\em Proceedings of the 47th Annual ACM Symposium on Theory of
  Computing (STOC)}, 2015.

\end{thebibliography}
